\documentclass[11pt]{article}

\usepackage[preprint]{acl}

\usepackage{times}
\usepackage{latexsym}

\usepackage{amsmath}
\usepackage{multirow}
\usepackage{graphicx}
\usepackage{makecell}
\usepackage{booktabs}
\usepackage{todonotes}
\usepackage{hyperref}
\usepackage{tikz}
\usepackage{pgfplots}
\usetikzlibrary{matrix}
\usepackage[normalem]{ulem}
\usepackage{xcolor}
\usepackage{natbib}

\usepackage{hyperref}       % hyperlinks
\usepackage{url}            % simple URL typesetting
\usepackage{booktabs}       % professional-quality tables
\usepackage{amsfonts}       % blackboard math symbols
\usepackage{nicefrac}       % compact symbols for 1/2, etc.
\usepackage{microtype}      % microtypography
\usepackage{xcolor}         % colors
\usepackage{caption}

\usepackage[T1]{fontenc}
\usepackage[utf8]{inputenc}

\usepackage{microtype}

\usepackage{inconsolata}

\usepackage{graphicx}

\title{On Membership Inference Attacks in Knowledge Distillation}

\author{Ziyao Cui\textsuperscript{*}, Minxing Zhang\textsuperscript{*}, Jian Pei \\
        Department of Computer Science\\
    Duke University\\
    Durham, NC 27705 \\
    \texttt{\{richard.cui, minxing.zhang, j.pei\}@duke.edu}}

\begin{document}
\maketitle

\begin{abstract}
Large language models (LLMs) are trained on massive corpora that may contain sensitive information, creating privacy risks under membership inference attacks (MIAs). Knowledge distillation is widely used to compress LLMs into smaller student models, but its privacy implications are poorly understood. We systematically evaluate how distillation affects MIA vulnerability across six teacher-student model pairs and six attack methods. 
% We find that distilled student models often match or exceed teacher models in MIA accuracy and can exhibit substantially higher member-specific attack success, indicating increased privacy risk. 
We find that distilled student models do not consistently exhibit lower MIA success than their teacher models, and in some cases demonstrate substantially higher member-specific attack success, challenging the assumption that knowledge distillation inherently improves privacy.
%We attribute this to mixed supervision in distillation: teacher model predictions align with ground truth on vulnerable examples, thereby student models learn overly confident behaviors that amplify separability between members and non-members. 
We attribute this to mixed supervision in distillation: for vulnerable training data points, teacher predictions often align with ground-truth labels, causing student models to learn overly confident predictions that amplify the separability between members and non-members; conversely, for non-vulnerable points, teacher predictions and ground truth frequently diverge, providing inconsistent learning signals.
To mitigate this, we propose three practical interventions -- restricting distillation to non-vulnerable points, adding a low-dimensional \textbf{Bottleneck Projection}, and a normalization variant (\textbf{NoNorm}). Experiments show these methods reduce both aggregate and member-specific MIA success while preserving model utility, improving privacy-utility trade-offs for distilled LLMs.\footnote{Our implementation and evaluation code are available at \url{https://github.com/richardcui18/mia-in-kd}.}

\begingroup
  \renewcommand\thefootnote{\fnsymbol{footnote}}% use symbol footnote
  \footnotetext[1]{Both authors contributed equally to this research.}
\endgroup

%  Large language models (LLMs) are trained on massive corpora that often contain sensitive or private information, creating privacy risks under membership inference attacks (MIAs). Knowledge distillation is widely used to compress LLMs into smaller, efficient student models, but the privacy consequences of distillation are not well understood. In this paper, we systematically evaluate how distillation affects vulnerability to MIA across six teacher–student model pairs and six representative attack methods. We show that distilled student models frequently match or exceed teacher models in aggregate MIA accuracy and can exhibit substantially higher member-specific attack success, indicating an increased risk to data exposure. We trace this behavior to the mixed supervision used in distillation: when the teacher model's soft predictions and the ground-truth signal align on vulnerable examples, the student model learns overly confident predictions that amplify separability between members and non-members. To mitigate this vulnerability, we introduce three practical interventions—distillation restricted to non-vulnerable inputs, a low-dimensional bottleneck projection, and a normalization variant (NoNorm)—and evaluate combinations of these techniques. Across experiments, these methods reduce member-specific MIA success while preserving model utility, yielding improved privacy–utility trade-offs for distilled LLMs. Our implementation and evaluation code are available at \url{https://anonymous.4open.science/r/mia-in-kd-C6D2}.
\end{abstract}

\section{Introduction}
\label{introduction}

Large Language Models (LLMs) have achieved remarkable success due to the scale and diversity of their pretraining data~\cite{liu2025datasets, zhang2025generalizability}. Moreover, as AI models become increasingly prevalent, privacy concerns have gained significant attention, with researchers investigating vulnerabilities and defenses across model architectures~\cite{carlini2021extracting, carlini2022membership,cui2025learning}. For LLMs, massive pretraining data introduces serious risks of training data exposure and privacy breaches~\cite{xie2024recall, wang2024recall}.
More specifically, the massive and heterogeneous nature of these datasets makes it infeasible to fully remove sensitive content, including copyrighted materials~\cite{meeus2024did, duarte2024cop} and personally identifiable information~\cite{mozes2023use, tang2023privacy}. Consequently, LLMs may memorize privacy-sensitive data, enabling attackers to infer training membership through Membership Inference Attacks (MIAs)~\cite{shokri2017membership}.

Prior work has extensively studied the detection of pretraining data in LLMs using MIAs~\cite{xie2024recall, yeom2018privacy, carlini2021extracting, carlini2022membership}. However, these studies largely analyze models in isolation. In parallel, model compression techniques such as knowledge distillation have been widely adopted to reduce the size and computational cost of modern LLMs, including Llama, Gemma, and BERT~\cite{sanh2019distilbert, jiao2019tinybert, lan2019albert, sun2020mobilebert, timiryasov2023baby, gemma2bdistilled}. Because distilled student models have fewer parameters and lower capacity, it is commonly assumed that distillation reduces memorization and improves privacy. This assumption, however, has not been systematically evaluated across diverse teacher-student pairs and MIA methods, leaving the privacy effects of distillation insufficiently understood.

To address this gap, as the first contribution in this paper, we evaluate six teacher-student model pairs across multiple architectures using six representative MIAs. We report three key findings. 
% First, distilled student models often achieve aggregate MIA accuracy comparable to or exceeding that of their teacher models. 
First, distilled student models do not consistently exhibit lower aggregate MIA accuracy than their teacher models.
Second, student models can exhibit higher member-specific attack accuracy even when their overall accuracy is lower, which represents a greater practical privacy risk~\cite{carlini2022membership}. 
% Third, we provide an explanation: the mixed supervision in distillation -- combining ground-truth labels and teacher predictions -- can reinforce memorization on vulnerable examples, leading the student model to produce overly confident outputs that amplify the separability between members and non-members.
Third, we provide an explanation: the mixed supervision in distillation -- combining ground-truth labels and teacher predictions -- can reinforce memorization on vulnerable training data points due to the alignment between teacher predictions and ground-truth, leading the student model to produce overly confident outputs that amplify the separability between members and non-members; on the other hand, for non-vulnerable data points, teacher predictions diverge from ground-truth, making the supervision inconsistent and failing to provide clear privacy benefits.

Motivated by this insight, as the second contribution, we propose and evaluate three targeted interventions to reduce membership leakage in distilled models. First, we introduce a data-selection strategy that restricts distillation to \emph{non-vulnerable} training data points, which reduces exposure to memorized points but may increase student perplexity due to reduced training data. To mitigate this utility loss, we propose two lightweight architectural modifications. The first is a low-dimensional 
\textbf{Bottleneck Projection} that limits representational capacity and discourages memorization. 
The second replaces layer normalization with \textbf{NoNorm}, a parameterized element-wise linear transformation. Across experiments, these interventions consistently reduce MIA success, and the architectural modifications in particular lower attack success without degrading model utility, yielding improved privacy-utility trade-offs compared to na\"ive distillation.

\paragraph{Outline.}
Section~\ref{sec:problem} formalizes membership inference and introduces the evaluation metrics and protocol. Section~\ref{sec:related} reviews related work on membership inference, knowledge distillation, and privacy-aware model compression. Section~\ref{sec:empirical} presents a systematic empirical analysis of membership inference vulnerability in teacher and student models and introduces diagnostics for privacy leakage in student models. Section~\ref{sec:methods} describes the proposed privacy-preserving distillation methods. Section~\ref{sec:experiments} reports experimental results, privacy-utility trade-offs, and additional analyses. Finally, Section~\ref{sec:conclusion} discusses limitations and directions for future work; supplementary material and extended ablations are provided in the Appendix.

\section{Problem Definition and Evaluation Metrics}
\label{sec:problem}

\subsection{Membership Inference Attacks}
\label{problem formulation: mia}

Consider a LLM $\mathcal{N}$ and a dataset $\mathcal{D}$ used to train $\mathcal{N}$. A \textbf{MIA method}~\cite{yeom2018privacy} $M$ takes a target data point $d$ as input and aims to determine whether $d\in \mathcal{D}$. Denote by $M(\mathcal{N},d)$ the attacker's prediction, where $M(\mathcal{N},d)=1$ if $M$ predicts $d \in \mathcal{N}$ and $0$ otherwise. In practice, $M$ may compute a confidence score $M'(\mathcal{N},d)$ and use a threshold $\tau$ to predict
the membership of $d$, that is,
%\begin{equation}
\(   M_\tau(\mathcal{N},d)=\mathbf{1}[M'(\mathcal{N},d)>\tau]
\),
%\end{equation}
where $\mathbf{1}$ is the indicator function.

\subsection{Knowledge Distillation}

In knowledge distillation, a teacher model $\mathcal{T}$ is used as a guiding framework to transfer knowledge to a student model $\mathcal{S}$, where the student model $\mathcal{S}$ is learned to mimic the performance of the teacher $\mathcal{T}$~\cite{shokri2017membership,xu2024survey}.%\todo{Should cite the original paper in addition to the survey.}. 

Let $\mathcal{D}_{\mathcal{T}}$ and $\mathcal{D}_{\mathcal{S}}$ denote the training datasets of the teacher model $\mathcal{T}$ and student model $\mathcal{S}$, respectively. We use the standard mixed-supervision distillation objective~\cite{hinton2015distilling}, in which the student model is trained to minimize a weighted sum of the supervised loss on ground-truth labels and a distillation loss that encourages the student model to match the teacher model’s predictive distribution. Specifically, for training data points $(x,y)\sim\mathcal{D}_{\mathcal{S}}$, the objective is
\begin{equation*}
    \begin{split}
        \mathcal{L}_{\text{distill}} = &\mathbb{E}_{(x,y)\sim\mathcal{D}_\mathcal{S}}\Big[\mathcal{L}_{\mathrm{CE}}\big(y,p_{\mathcal{S}}(\cdot\mid x)\big) \\
        &+ \lambda\,\mathrm{KL}\big(p_{\mathcal{T}}(\cdot\mid x)\,\|\,p_{\mathcal{S}}(\cdot\mid x)\big)\Big]
    \end{split}
\end{equation*}
where $p_{\mathcal{T}}(\cdot \mid x)$ and $p_{\mathcal{S}}(\cdot \mid x)$ are the predictive distributions of the teacher and student models for training data (prefix) $x$, $\mathcal{L}_{\mathrm{CE}}$ denotes cross-entropy loss, $\mathrm{KL}(\cdot\|\cdot)$ is the Kullback-Leibler divergence, and $\lambda \ge 0$ controls the trade-off between the two terms.
%Since knowledge is transferred from $\mathcal{T}$ to $\mathcal{S}$ during knowledge distillation, Given a data point $d\in \mathcal{D}_t$, since the teacher model $\mathcal{T}$ is trained using $d$ and knowledge from $\mathcal{T}$ is transferred to $\mathcal{S}$ during knowledge distillation, we consider $d$ to also be used for training $\cal S$, thus $d\in \mathcal{D}_s$. Therefore, $\mathcal{D}_t \subseteq \mathcal{D}_s$. It follows that $\mathcal{D}_t$ is the training dataset for both $\mathcal{T}$ and $\mathcal{S}$, and we denote $\mathcal{D}_t=\mathcal{D}$ for simplicity. Moreover, let  $\mathcal{D}'$ be a dataset that was not used to train $\mathcal{T}$ or $\mathcal{S}$. $\mathcal{D}'$ can be obtained using several methods, such as selecting a dataset that was released after $\mathcal{T}$ and $\mathcal{S}$. 

\subsection{Problem Definition}\label{sec:prob-defn}

Consider a model $\mathcal{T}$ trained on dataset $\mathcal{D}_{\mathcal{T}}$. Let $\mathcal{D}\subseteq \mathcal{D}_{\mathcal{T}}$ be a set of member data points and let $\mathcal{D}'$ be a non-member set such that $\mathcal{D}' \cap \mathcal{D}_{\mathcal{T}} = \emptyset$. Given a MIA method $M_{\tau}$ with threshold $\tau$, we define the \textbf{MIA true positive rate} as
$
TPR(\mathcal{T}) = \frac{1}{|\mathcal{D}|} \sum_{x \in \mathcal{D}} \mathbf{1}[M_{\tau}(\mathcal{T}, x) = 1]
$,
and the \textbf{MIA true negative rate} as
$
TNR(\mathcal{T}) = \frac{1}{|\mathcal{D}'|} \sum_{x' \in \mathcal{D}'} \mathbf{1}[M_{\tau}(\mathcal{T}, x') = 0]
$.
We define the \textbf{MIA accuracy} on model $\mathcal{T}$ as the average of these two quantities:
\begin{equation}
\label{eq: teacher accuracy definition}
A(\mathcal{T}) = \frac{1}{2}\big[TPR(\mathcal{T}) + TNR(\mathcal{T})\big].
\end{equation}

The central question studied in this paper is whether distilled student models $\mathcal{S}$ are more vulnerable to MIAs than their teacher models $\mathcal{T}$. We assess vulnerability using both aggregate MIA accuracy, $A(\mathcal{T})$ versus $A(\mathcal{S})$, and member-specific MIA accuracy, $TPR(\mathcal{T})$ versus $TPR(\mathcal{S})$, and investigate how distillation can be modified to improve student robustness to membership inference.

\section{Related Work}
\label{sec:related}

\subsection{MIA Methods}

MIAs, introduced by~\citet{shokri2017membership}, aim to determine whether a data point was used to train a model. While MIAs have been studied in settings such as diffusion models~\cite{carlini2023extracting} and multi-layer perceptrons~\cite{watson2021importance}, applying MIAs to LLMs presents distinct challenges. LLM training data are typically not public~\cite{touvron2023llama, anthropic2024claude, guo2025deepseek}, complicating evaluation due to missing ground-truth membership labels, and modern LLMs are often trained for a single epoch over massive corpora, reducing classical memorization signals~\cite{carlini2023quantifying, shi2023detecting}. Despite these challenges, several MIA methods have been developed for LLMs, including loss-based attacks~\cite{yeom2018privacy}, zlib-normalized loss~\cite{carlini2021extracting}, reference-model attacks using likelihood ratio tests~\cite{carlini2022membership}, and ReCaLL, which leverages relative conditional log-likelihoods~\cite{xie2024recall}.

However, existing studies primarily analyze MIAs on individual models in isolation and do not consider how vulnerability changes across related models, such as teacher-student pairs produced by knowledge distillation. In particular, prior MIA methods~\cite{yeom2018privacy, carlini2021extracting} were not designed to assess privacy trade-offs introduced by distillation. \emph{Our work addresses this gap by systematically evaluating MIA behavior across distilled teacher-student model pairs and using these insights to develop privacy-preserving distillation methods}.

\subsection{Knowledge Distillation}

Knowledge distillation~\cite{hinton2015distilling, buciluǎ2006model} is a model compression technique in which a smaller student model is trained to mimic a larger teacher model using a combination of supervised loss and a distillation loss that aligns predictions by student and teacher models~\cite{gou2021knowledge, xu2024survey}. Distillation has been widely applied to LLMs, including DistilBERT~\cite{sanh2019distilbert} and subsequent methods that focus on matching output distributions~\cite{song2020lightpaff, liang2020mixkd, zhang2023not}. Other approaches exploit intermediate representations as training signals, enabling the student model to imitate the teacher model’s hidden states through layer-wise distillation~\cite{romero2014fitnets, sun2019patient, jiao2019tinybert}.

Prior work on knowledge distillation has largely emphasized efficiency and performance, with limited attention to privacy. In contrast, \emph{our work studies knowledge distillation through the lens of MIAs, analyzing how the distillation process alters privacy risk across teacher-student model pairs}. Rather than treating distillation as inherently privacy-preserving, we leverage the teacher model’s MIA vulnerability as a signal to understand and mitigate privacy leakage in the student model. As distillation becomes a standard component of LLM deployment, incorporating privacy-aware objectives is essential for responsible model compression.

\subsection{Privacy in Knowledge Distillation}

\begin{table*}[t]
  \centering
  \begin{minipage}[c]{\textwidth}
    \centering
    \scriptsize
%    \vspace{0.5em}
    \begin{tabular}{ll|cccccc}
    \toprule
    Teacher Model & Student Model & ReCall & Loss & Zlib & Mink & Mink++ & Ref Model \\
    \midrule
    \multirow{1}{*}{Pythia} & DistilPythia & \textbf{0.555} / 0.565 & \textbf{0.442} / 0.444 & \textbf{0.613} / 0.635 & \textbf{0.316} / 0.414 & 0.501 / 0.501 & 0.648 / \textbf{0.579} \\
    \midrule
    \multirow{3}{*}{Gemma 2 27B} & Gemma 2 2B & 0.667 / 0.667 & \textbf{0.494} / 0.525 & 0.556 / \textbf{0.481} & \textbf{0.543} / 0.556 & 0.537 / \textbf{0.451} & 0.494 / \textbf{0.420} \\
     & Gemma 2 2B Distilled & 0.667 / \textbf{0.494} & \textbf{0.494} / 0.537 & 0.556 / 0.556 & \textbf{0.543} / 0.580 & 0.537 / \textbf{0.432} & 0.494 / \textbf{0.426} \\
     & Gemma 2 9B & \textbf{0.667} / 0.704 & 0.494 / \textbf{0.475} & 0.556 / \textbf{0.531} & 0.543 / 0.543 & 0.537 / \textbf{0.481} & 0.494 / \textbf{0.426} \\
     \midrule
    \multirow{2}{*}{Llama 3.1 8B} & Llama 3.2 1B & 0.702 / \textbf{0.682} & \textbf{0.303} / 0.311 & 0.552 / \textbf{0.532} & \textbf{0.300} / 0.309 & \textbf{0.311} / 0.335 & 0.441 / \textbf{0.349} \\
     & Llama 3.2 3B & \textbf{0.702} / 0.806 & \textbf{0.303} / 0.314 & 0.552 / \textbf{0.538} & \textbf{0.300} / 0.311 & 0.311 / \textbf{0.205} & 0.441 / \textbf{0.381} \\
    \bottomrule
    \end{tabular}
    \captionof{table}{Comparison of aggregate MIA accuracy for teacher-student model pairs. Each cell reports teacher / student MIA accuracy, with \textbf{bold} indicating the lower (more privacy-preserving) value for each pair and attack method.}
    \label{table:teacher_student_comparison}  \end{minipage}
\end{table*}

A growing body of work has studied the privacy implications of knowledge transfer mechanisms such as knowledge distillation. Several approaches propose distillation-based defenses against membership leakage, including ensemble and self-distillation schemes and cross-distillation protocols~\cite{tang2022selfdistillation, chourasia2022crossdistill, shejwalkar2021knowledge, zheng2021resisting}. These methods, however, focus on specific engineered variants rather than the canonical mixed-supervision distillation pipeline commonly used in practice. In contrast, \emph{our work systematically evaluates this standard pipeline across diverse LLM families and identifies conditions under which distillation amplifies membership leakage}.

Related empirical studies have also questioned the privacy benefits of distillation. For example,~\citet{jagielski2023students} show that student models may replicate teacher model behaviors under certain conditions, yielding limited privacy gains. Their setting, however, considers distillation solely from teacher model outputs, without access to ground-truth labels, whereas \emph{our study analyzes the more prevalent mixed-supervision objective~\cite{hinton2015distilling} that combines teacher model predictions and labeled data}. Earlier work by~\citet{jagannatha2021membership} reports lower privacy leakage in DistilBERT compared to BERT, but is limited to a single model pair and a narrow clinical dataset. \emph{Our work expands beyond these settings by examining multiple teacher-student model pairs and datasets, providing a broader understanding of how distillation affects privacy and how it can be modified to improve robustness against membership inference}.

\section{MIAs in Distilled LLMs}
\label{sec:empirical}

We present a systematic empirical analysis of membership inference vulnerability under knowledge distillation, evaluating six teacher-student pairs across multiple architectures using six representative attacks and the metrics defined in Section~\ref{sec:prob-defn}.

\subsection{Experimental Setup}
\label{experimental setup}

% \begin{figure}[t]
%   \centering
%   \begin{minipage}[c]{0.48\textwidth}
%     \centering
%     \includegraphics[width=0.7\linewidth]{figures/next_token_prob.png}
%     \captionof{figure}{Average probability assigned by the teacher model to the ground-truth next token, by membership inference vulnerability on the teacher. Vulnerable (successful attack) examples exhibit higher probabilities, indicating stronger alignment between teacher and ground-truth learning signals.}
%     \label{fig:next_token_prob}
%   \end{minipage}
%     \hfill
%   \begin{minipage}[c]{0.48\textwidth}
%     \centering
%     \includegraphics[width=0.7\linewidth]{figures/kl_div.png}
%     \captionof{figure}{KL divergence between the teacher's predictive distribution and the ground-truth distribution, by membership inference vulnerability on the teacher. Vulnerable (successful attack) examples show lower divergence, reflecting greater agreement between teacher and ground-truth signals during distillation.}
%     \label{fig:kl_div}
%   \end{minipage}
% \end{figure}

\paragraph{Models and Datasets.}

We evaluate three model families with known training data and publicly available distilled variants. \textbf{Pythia}~\cite{biderman2023pythia} is trained on the ArXiv subset of the Pile~\cite{gao2020pile}, with DistilPythia~\cite{distilpythia} as the student model. \textbf{Gemma 2 27B}~\cite{team2024gemma} is trained on the WikiMIA dataset~\cite{shi2023detecting}, with Gemma 2 9B, Gemma 2 2B, and Gemma 2 2B distilled~\cite{team2024gemma, gemma2bdistilled} as student models. \textbf{Llama 3.1 8B}~\cite{grattafiori2024llama} is trained on the ArXiv subset of the Pile~\cite{gao2020pile}, with Llama 3.2 3B and Llama 3.2 1B~\cite{grattafiori2024llama} as student models.

For non-member evaluation, we use the built-in non-member split of WikiMIA for Gemma models~\cite{shi2023detecting}. For all other models, we use the WebInstructSub-prometheus dataset~\cite{WebInstructSub-prometheus}, released in May 2024, after the model release dates, ensuring it was not used during training -- a method widely adopted by existing works~\cite{meeus2025sok,xie2024recall,wang2024recall}.

\paragraph{MIA Methods.}

We evaluate membership inference using six representative attacks. \textbf{ReCaLL}~\cite{xie2024recall} measures changes in conditional log-likelihood when prefixing inputs with non-member context. \textbf{Loss}~\cite{yeom2018privacy} uses the per-example loss as a membership score, while \textbf{Zlib}~\cite{carlini2021extracting} normalizes this loss by zlib compression entropy. The \textbf{Reference-model} attack extends the loss-based approach by training shadow models with and without the target data point and performing a likelihood-ratio test. \textbf{Min-K\%}~\cite{shi2023detecting} computes membership scores from the average log-likelihood of the lowest-probability $k\%$ tokens, and \textbf{Min-K\%++} further calibrates these scores using the mean and standard deviation over the vocabulary.

We leverage cross-validation to select hyperparameters for these attacks: for each dataset and target model, we tune attack-specific hyperparameters -- such as the prefix length for ReCaLL and the value of $k$ for Min-K\% and Min-K\%++ -- to maximize average attack performance. We report detailed hyperparameters for each scenario in Appendix~\ref{app:hyperparam}.

As described in Section~\ref{sec:problem}, each method outputs a confidence score per data point, which is converted to a binary prediction using a threshold selected via ROC analysis to maximize MIA accuracy while ensuring balanced predicted classes. All experiments are conducted on NVIDIA A5000/A6000 GPUs.

\subsection{Aggregate MIA Accuracy}
\label{sec:aggregate}

\begin{table*}[t]
  \centering
  \begin{minipage}[c]{\textwidth}
    \centering
    \scriptsize
%    \vspace{0.5em}
    \begin{tabular}{ll|cccccc}
    \toprule
    Teacher Model & Student Model & ReCall & Loss & Zlib & Mink & Mink++ & Ref Model \\
    \midrule
    \multirow{1}{*}{Pythia} & DistilPythia & \textbf{0.658} / 0.673 & 0.631 / \textbf{0.630} & \textbf{0.234} / 0.290 & \textbf{0.444} / 0.683 & 0.993 / 0.993 & \textbf{0.620} / 0.636 \\
    \midrule
    \multirow{3}{*}{Gemma 2 27B} & Gemma 2 2B & \textbf{0.615} / 0.670 & \textbf{0.606} / 0.661 & 0.156 / \textbf{0.037} & 0.761 / \textbf{0.495} & 0.688 / \textbf{0.404} & 0.780 / \textbf{0.376} \\
     & Gemma 2 2B Distilled & 0.615 / \textbf{0.532} & \textbf{0.606} / 0.716 & 0.156 / \textbf{0.101} & 0.761 / \textbf{0.376} & \textbf{0.688} / 0.771 & 0.780 / \textbf{0.358} \\
     & Gemma 2 9B & 0.615 / \textbf{0.459} & 0.606 / \textbf{0.505} & \textbf{0.156} / 0.165 & 0.761 / \textbf{0.633} & 0.688 / \textbf{0.450} & 0.780 / \textbf{0.339} \\
    \midrule
    \multirow{2}{*}{Llama 3.1 8B} & Llama 3.2 1B & 0.968 / \textbf{0.937} & \textbf{0.451} / 0.513 & \textbf{0.497} / 0.567 & \textbf{0.447} / 0.508 & \textbf{0.486} / 0.604 & 0.563 / 0.563 \\
     & Llama 3.2 3B & \textbf{0.968} / 0.974 & \textbf{0.451} / 0.510 & \textbf{0.497} / 0.515 & \textbf{0.447} / 0.502 & 0.486 / \textbf{0.320} & \textbf{0.563} / 0.563 \\
    \bottomrule
    \end{tabular}
    \captionof{table}{Comparison of member-specific MIA accuracy for teacher-student model pairs. Each cell reports teacher / student member-specific MIA accuracy, with \textbf{bold} indicating the lower (more privacy-preserving) value for each pair and attack method.}
    \label{table:member_specific_comparison}  
    \end{minipage}
\end{table*}

Table~\ref{table:teacher_student_comparison} reports aggregate MIA accuracy for teacher-student model pairs. Across six model pairs and six attack methods, student models frequently achieve MIA accuracy comparable to, and in several cases exceeding, that of their teachers. Specifically, the teacher model exhibits lower accuracy in 15 cases, the student model in 17 cases, and both models tie in 4 cases. 

%\paragraph{Significance Test} 

We further test whether teacher models consistently exhibit higher MIA accuracy than their distilled student models using a one-sided sign test, with the null hypothesis
$H_0:\; P(A(\mathcal{T}) > A(\mathcal{S})) \leq 0.5
$ and alternative hypothesis $H_1:\; P(A(\mathcal{T}) > A(\mathcal{S})) > 0.5$, where $A(\mathcal{T})$ and $A(\mathcal{S})$ denote the teacher and student MIA accuracies, respectively.
Across all comparisons, the teacher model exhibits higher MIA accuracy in 17 cases, yielding a $p$-value of 0.43. We therefore fail to reject $H_0$, indicating that \emph{knowledge distillation does \textbf{not} consistently reduce membership inference risk}. This result challenges the common assumption that model compression alone improves privacy.

\subsection{Member-Specific MIA Accuracy}

Since member-specific privacy risks are more practically significant than aggregate metrics~\cite{carlini2022membership}, we explore the member-specific MIA accuracy for teacher-student model pairs.

As illustrated in Table~\ref{table:member_specific_comparison}, the teacher model exhibits lower accuracy in 18 cases, the student model in 16 cases, and both models tie in 2 cases. 
For Pythia and Llama families, the student model member-specific MIA accuracy increases by 9.81\% and 2.29\%, respectively, compared to the teacher model.
Notably, under the reference-model attack for the Pythia family, the student model achieves lower aggregate accuracy than the teacher model (0.579 vs.\ 0.648; Table~\ref{table:teacher_student_comparison}) while exhibiting higher member-specific accuracy (0.636 vs.\ 0.620; Table~\ref{table:member_specific_comparison}). This demonstrates that aggregate MIA accuracy can obscure increased member-specific vulnerability and that lower overall attack success does not necessarily imply stronger privacy protection. 

We also conduct the one-sided sign test with a similar setting as Section~\ref{sec:aggregate}. Across all comparisons, the teacher model exhibits higher MIA accuracy in 16 cases, giving a $p$-value of 0.432. We again fail to reject $H_0$, indicating that knowledge distillation also does not consistently reduce member-specific risk.

\subsection{Why May Student Models Exhibit Greater Privacy Leakage?}

Teacher and student models differ fundamentally in their training signals. Teacher models are trained solely with ground-truth supervision, whereas student models optimize a mixed objective that combines the ground-truth loss with a distillation loss, typically the KL divergence between teacher model and student model predictions. Consequently, student models learn from two supervision signals: ground-truth labels and the teacher model’s soft predictions.

These signals are not equally aligned across training data points. For training data points that are already vulnerable to MIAs (i.e., MIA on this point is successful) on the teacher model, the teacher model assigns high probability to the ground-truth token and exhibits low divergence from the ground-truth distribution. In this case, distillation reinforces supervised learning, allowing the student model to fit these data points especially well. In contrast, for less vulnerable training data points (i.e., MIA is unsuccessful), teacher model predictions are less confident and deviate more from the ground-truth, introducing noise into the distillation signal and failing to provide clear privacy benefits.

% \begin{figure}[t]
%   \centering
%   \begin{minipage}[c]{0.5\textwidth}
%     \centering
%     \includegraphics[width=\linewidth]{figures/next_token_prob.png}
%   \end{minipage}

%   % \vspace{0.5em}

%   \begin{minipage}[c]{0.5\textwidth}
%     \centering
%     \includegraphics[width=\linewidth]{figures/kl_div.png}
%   \end{minipage}

%   \caption{Teacher \& ground-truth alignment stratified by membership inference vulnerability. 
%   \textbf{Top:} Average probability assigned by the teacher model to the ground-truth next token. 
%   \textbf{Bottom:} KL divergence between the teacher model's predictive distribution and the ground-truth distribution. 
%   Vulnerable training data points exhibit higher probabilities and lower divergence, indicating stronger alignment between the two signals.}
%   \label{fig:teacher_alignment}
% \end{figure}

\begin{figure}[t]
  \centering
  \begin{minipage}[c]{0.48\columnwidth}
    \centering
    \includegraphics[width=\linewidth]{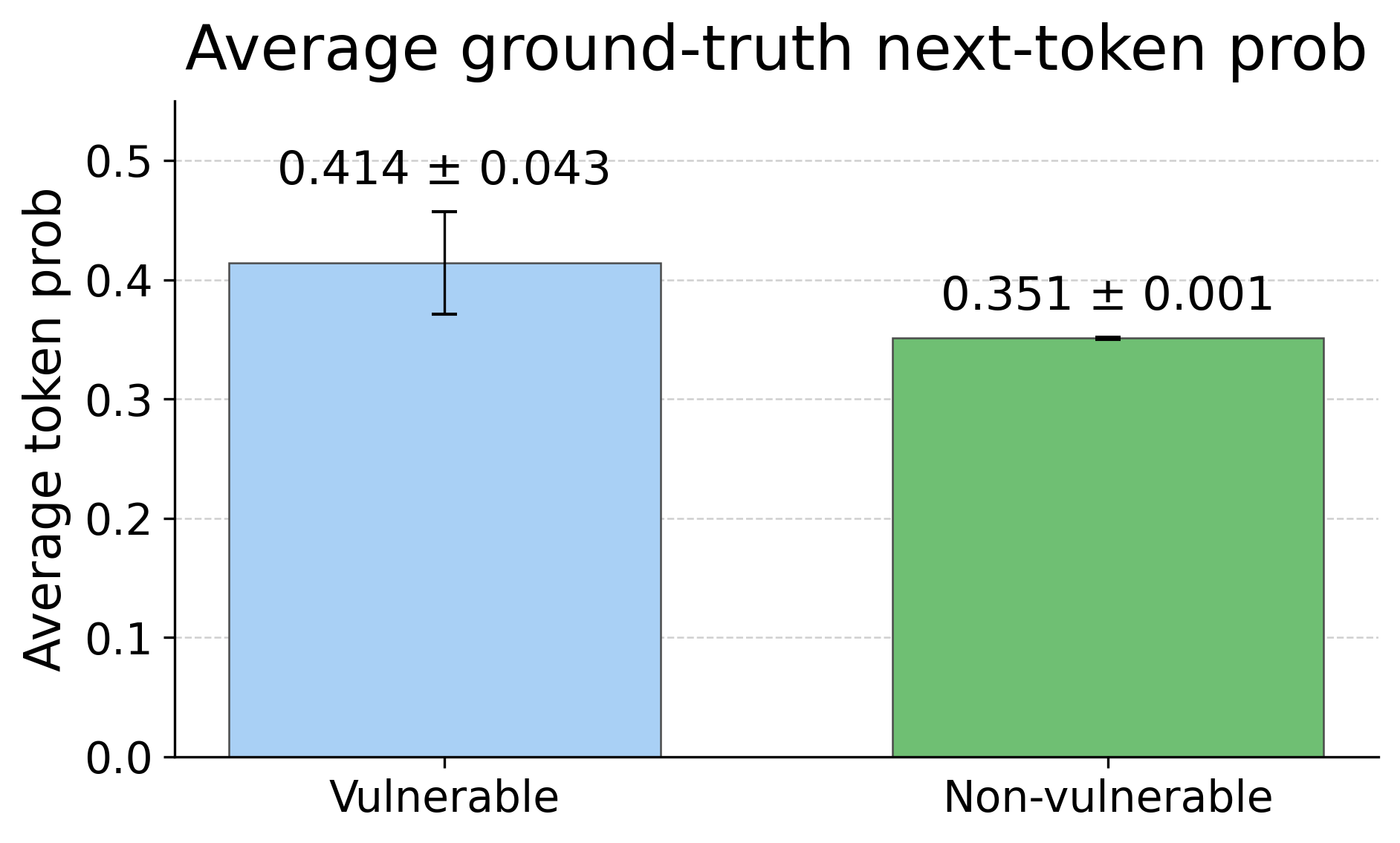}
  \end{minipage}%
  \hfill%
  \begin{minipage}[c]{0.48\columnwidth}
    \centering
    \includegraphics[width=\linewidth]{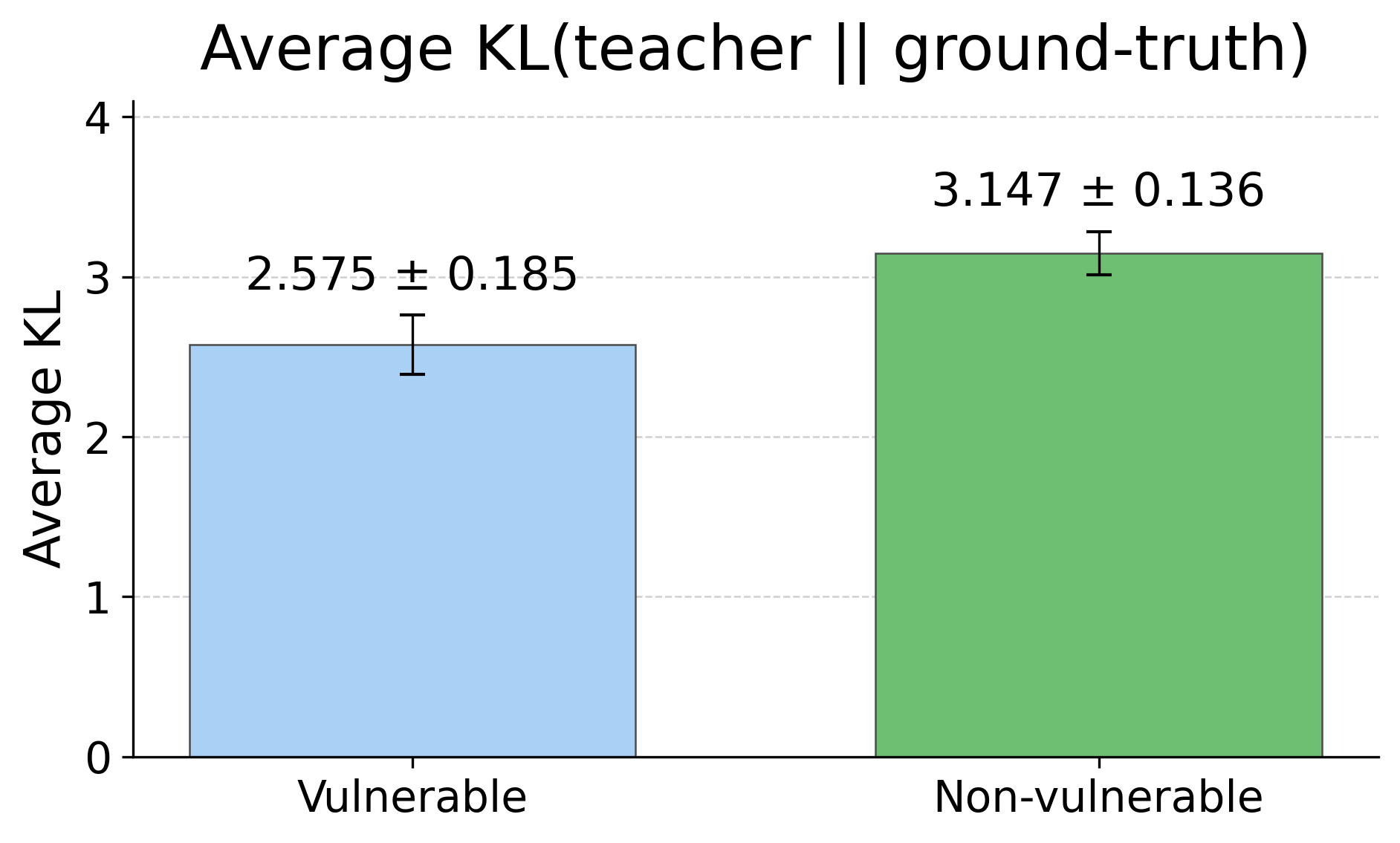}
  \end{minipage}
  \caption{Teacher \& ground-truth alignment stratified by membership inference vulnerability. 
  \textbf{Left:} Average probability assigned by the teacher model to the ground-truth next token. 
  \textbf{Right:} KL divergence between the teacher model's predictive distribution and the ground-truth distribution.}
  \label{fig:teacher_alignment}
\end{figure}

We quantify this effect using two metrics: the teacher model’s probability assigned to the ground-truth next token and the KL divergence between the teacher model’s predictive distribution and the ground-truth distribution. As shown in Figure~\ref{fig:teacher_alignment}, attack-vulnerable training data points exhibit higher next-token probabilities and lower KL divergence, indicating strong alignment between supervision signals, whereas less vulnerable data points show the opposite trend. This asymmetry leads student models to preferentially learn and overfit already vulnerable points, amplifying the confidence gap between members and non-members and strengthening MIA decision boundaries.

\section{Privacy-Preserving Distillation Methods}
\label{sec:methods}

In this section, we develop methods for reducing membership inference risk in distilled models. We first consider a simple data-selection method that restricts distillation to \emph{non-vulnerable} training data points. Although this approach directly reduces exposure to memorized data points, it also weakens the training signal and can degrade model utility. Motivated by this trade-off, we introduce two lightweight architectural modifications -- \textbf{bottleneck projection} and \textbf{NoNorm} -- that can be applied during distillation to reduce memorization while largely preserving utility.

\subsection{Distillation on Non-Vulnerable Data}
\label{sec:nonvulnerable}

We consider a simple data-selection strategy that uses privacy signals from the teacher model to limit student model exposure to memorized content. We apply a MIA to the teacher model $\mathcal{T}$ on its training dataset $\mathcal{D}$ and partition $\mathcal{D}$ into two disjoint subsets: a \emph{vulnerable} set
$
\mathcal{D}_v = \{x \in \mathcal{D} \mid M(\mathcal{T}, x) = 1\}
$,
containing data points identified as members by the attack, and a \emph{non-vulnerable} set
$
\mathcal{D}_{nv} = \{x \in \mathcal{D} \mid M(\mathcal{T}, x) = 0\}
$.
By construction, $\mathcal{D}_v \cup \mathcal{D}_{nv} = \mathcal{D}$ and $\mathcal{D}_v \cap \mathcal{D}_{nv} = \emptyset$.

This partition captures privacy-relevant structure in the teacher model: data points in $\mathcal{D}_v$ exhibit behaviors that are easily distinguishable from non-members, whereas those in $\mathcal{D}_{nv}$ do not. Accordingly, a natural approach is to restrict distillation to $\mathcal{D}_{nv}$, which directly reduces exposure to highly memorized data points and limits the transfer of membership signals from teacher to student.

\paragraph{Empirical Evidence of Effectiveness.}

\begin{table}[ht]
\centering
\tiny
\begin{tabular}{l|cccccc}
\toprule
Model & ReCaLL & Loss & Zlib & Min-K & Min-K++ & Ref \\
\midrule
\multicolumn{7}{c}{\textbf{Evaluated on Vulnerable ($\mathcal{D}_v$)}} \\
\midrule
Non-Vulnerable & \textbf{0.250} & \textbf{0.205} & \textbf{0.114} & \textbf{0.364} & \textbf{0.318} & \textbf{0.114} \\
Full     & 1.000 & 1.000 & 0.977 & 0.818 & 0.864 & 1.000 \\
\midrule
\multicolumn{7}{c}{\textbf{Evaluated on Non-vulnerable ($\mathcal{D}_{nv}$)}} \\
\midrule
Non-Vulnerable & 1.000 & 0.976 & 0.976 & 0.952 & 0.929 & 1.000 \\
Full     & 1.000 & \textbf{0.952} & \textbf{0.452} & \textbf{0.833} & \textbf{0.905} & 1.000 \\
\midrule
\multicolumn{7}{c}{\textbf{Evaluated on Member ($\mathcal{D}$)}} \\
\midrule
Non-Vulnerable & \textbf{0.593} & \textbf{0.556} & \textbf{0.432} & \textbf{0.630} & \textbf{0.593} & \textbf{0.519} \\
Full     & 1.000 & 0.975 & 0.556 & 0.827 & 0.889 & 1.000 \\
\midrule
\multicolumn{7}{c}{\textbf{Evaluated on Non-member ($\mathcal{D}'$)}} \\
\midrule
Non-Vulnerable & 0.963 & 0.988 & \textbf{0.988} & \textbf{0.877} & 0.914 & 1.000 \\
Full     & 0.963 & \textbf{0.975} & 1.000 & 0.914 & \textbf{0.877} & 1.000 \\
\bottomrule
\end{tabular}
\caption{MIA accuracies for DistilPythia student models trained using only non-vulnerable data points vs. trained on the full dataset. \textbf{Bold} entries indicate lower MIA accuracy (stronger privacy).}
\label{tab:non-vul-accuracy}
\end{table}

\begin{table}[t]
\centering
\scriptsize
\begin{tabular}{l|c c c}
\toprule
Model & Vulnerable & Non-vulnerable & Members \\
\midrule
Non-Vulnerable & 972.96 & 188.06 & 436.01 \\
Full     & \textbf{94.45} & \textbf{149.54} & \textbf{118.21} \\
\bottomrule
\end{tabular}
\caption{Perplexity of student models evaluated on the Vulnerable, Non-vulnerable, and Member subsets. \textbf{Bold} entries indicate lower perplexity (better model utility).}
\label{tab:non-vul-perplexity}
\end{table}

We evaluate the non-vulnerable-only distillation strategy using DistilPythia derived from a Pythia teacher. Table~\ref{tab:non-vul-accuracy} reports MIA accuracies for six attacks (ReCaLL, Loss, Zlib, Min-K\%, Min-K\%++, and Reference-model), stratified by four subsets: \emph{Vulnerable} ($\mathcal{D}_v$), \emph{Non-vulnerable} ($\mathcal{D}_{nv}$), \emph{Member} ($\mathcal{D}$), and \emph{Non-member} ($\mathcal{D}'$). We compare two student model variants: distillation using only $\mathcal{D}_{nv}$ (``Non-Vulnerable'') and distillation using the full dataset $\mathcal{D}$ (``Full'').

Restricting distillation to non-vulnerable data points substantially reduces attack success on both vulnerable and member subsets. Averaged across attacks, the non-vulnerable-only student reduces MIA accuracy by 75.02\% on $\mathcal{D}_v$ and 35.20\% on $\mathcal{D}$, while achieving comparable accuracy on non-member data. These results confirm the privacy benefit of excluding highly memorized training data points during distillation.

However, this approach incurs clear utility costs. Table~\ref{tab:non-vul-perplexity} reports perplexity on vulnerable, non-vulnerable, and member subsets, showing that the non-vulnerable-only student model consistently exhibits higher perplexity, reflecting degraded language-model quality due to reduced training data.

Overall, while simple and effective at reducing membership leakage, non-vulnerable-only distillation substantially weakens the training signal by reducing the effective training dataset size. With this, it demonstrates that privacy leakage can be mitigated through data selection, while also highlighting the need for complementary methods that reduce leakage without incurring significant utility loss. This motivates the architectural interventions introduced next.

\begin{table*}[t]
\centering
\scriptsize
\begin{tabular}{l|cccccc}
\toprule
Model & ReCaLL & Loss & Zlib & Min-K & Min-K++ & Ref \\
\midrule
\multicolumn{7}{c}{\textbf{Evaluated on Member ($\mathcal{D}$)}} \\
\midrule
None
 & 1.000 & 1.000 & 0.802 & 1.000 & 0.951 & 1.000 \\
Bottleneck Projection
 & \textbf{0.802} & \textbf{0.951} & \textbf{0.667} & \textbf{0.926} & 0.901 & \textbf{0.975} \\
NoNorm
 & 0.988 & 1.000 & 0.728 & \textbf{0.926} & \textbf{0.852} & 1.000 \\
All
 & 1.000 & 1.000 & \textbf{0.667} & 0.938 & 0.914 & 1.000 \\
\midrule
\multicolumn{7}{c}{\textbf{Evaluated on Non-member ($\mathcal{D}'$)}} \\
\midrule
None
 & 0.963 & 0.988 & 1.000 & \textbf{0.926} & 0.963 & 1.000 \\
Bottleneck Projection
 & \textbf{0.926} & \textbf{0.975} & 1.000 & 0.951 & 0.926 & 1.000 \\
NoNorm
 & 1.000 & 0.988 & 1.000 & 0.988 & 0.988 & 1.000 \\
All
 & 1.000 & 0.988 & 1.000 & 0.963 & \textbf{0.914} & 1.000 \\
\bottomrule
\end{tabular}
\caption{MIA accuracies for DistilPythia student models under proposed architectural improvements. \textbf{Bolded} entries denote the lower MIA accuracy (better privacy).}
\label{tab:bottleneck-nonorm-acc}
\end{table*}

\subsection{Bottleneck Projection}
\label{sec:bottleneck}

Motivated by evidence that larger models are more prone to memorization and MIAs~\cite{carlini2021extracting}, we introduce a simple architectural modification that limits representational capacity: a low-dimensional \textbf{Bottleneck Projection} in the feed-forward network. Instead of the standard single projection from hidden dimension $H$ to intermediate size $I$ (typically $I \approx 4H$), we first project hidden states into a compact bottleneck space of dimension $B \ll H$, followed by expansion to $I$. This two-step projection has parameter cost $O(HB + BI)$, compared to $O(HI)$ for the standard design, yielding substantial parameter and computational savings when $B \ll I$.

Beyond efficiency, the \textbf{Bottleneck Projection} constrains intermediate representations, limiting the model’s ability to encode fine-grained, training data-specific signals. This restriction reduces memorization and weakens features exploited by MIAs. The \textbf{Bottleneck Projection} method is a lightweight modification that integrates seamlessly into existing transformer blocks and can be applied only to student models, preserving the capacity of the teacher.

\subsection{NoNorm}

% \citet{sun2020mobilebert} showed that replacing layer normalization can simplify model architecture and reduce inference latency\footnote{Although NoNorm has been proposed primarily to reduce model latency, it has not been studied as a defense mechanism against MIAs for privacy preservation in the LLM knowledge distillation scenario}. 
Motivated by \citet{sun2020mobilebert}, who showed that replacing layer normalization can simplify model architecture and reduce inference latency\footnote{Although NoNorm has been proposed primarily to reduce model latency, it has not been studied as a defense mechanism against MIAs for privacy preservation in the LLM knowledge distillation scenario.}, we introduce \textbf{NoNorm} as a privacy-preserving technique for knowledge distillation in LLMs.
More specifically, the mean and variance computations in layer normalization may inadvertently encode information about the training data, posing a potential privacy risk. To address this concern, we replace layer normalization with a simpler element-wise affine transformation:
\[
\text{NoNorm}(\mathbf{h}) = \boldsymbol{\gamma} \circ \mathbf{h} + \boldsymbol{\beta},
\]
where $\boldsymbol{\gamma}, \boldsymbol{\beta} \in \mathbb{R}^n$, $n$ is the number of channels, $\mathbf{h}$ denotes the hidden state, and $\circ$ is the Hadamard product.

This modification has two key advantages. First, \textbf{NoNorm} avoids computing training data statistics, eliminating a potential channel through which training data information could be memorized and exploited by MIAs. Second, the transformation in \textbf{NoNorm} improves inference efficiency, yielding a simpler model that is less prone to memorization and thus potentially more robust to MIAs.

\section{Experimental Evaluation of Privacy-Preserving Distillation Methods}
\label{sec:experiments}

We evaluate the effectiveness of the proposed architectural interventions -- \textbf{Bottleneck Projection} and \textbf{NoNorm} -- in reducing MIA vulnerability in distilled models. 

\subsection{Setup and Baselines}

All experiments use DistilPythia student models trained for 30 epochs.\footnote{We focus on the Pythia family due to computational constraints.} When applied, the \textbf{Bottleneck Projection} uses dimensionality $B=384$, reducing the intermediate representation by a factor of two relative to the standard setting and providing a favorable privacy-utility trade-off, as validated by the ablation study in Appendix~\ref{app:ablation-bottleneck}.

We compare four student model variants derived from the same teacher: (i) \textbf{None}, a model trained with standard distillation with no privacy protection; (ii) \textbf{Bottleneck Projection}, which incorporates a low-dimensional \textbf{Bottleneck Projection} ($B=384$) in the feed-forward layers; (iii) \textbf{NoNorm}, which replaces layer normalization with an element-wise affine transformation; and (iv) \textbf{All}, which combines \textbf{Bottleneck Projection} and \textbf{NoNorm}. Each variant is evaluated using six representative MIAs, with results reported separately for member and non-member data.

\subsection{MIA Accuracies}

Table~\ref{tab:bottleneck-nonorm-acc} reports MIA accuracy for all student model variants. On member data, the model with no privacy protection (\textbf{None}) exhibits the highest attack success across all methods. Introducing \textbf{Bottleneck Projection}, \textbf{NoNorm}, and their combination (\textbf{All}) reduces aggregate member-side attack accuracy by 9.45\%, 4.71\%, and 4.49\%, respectively, indicating that both interventions effectively suppress memorization signals exploited by MIAs.

On non-member data, attack accuracies remain broadly comparable across variants. Relative to the model with no privacy protection, \textbf{Bottleneck Projection} slightly reduces non-member accuracy by 1.05\%, while \textbf{NoNorm} and the combined model (\textbf{All}) increase it by 2.19\% and 0.46\%, respectively. These changes are small compared to the gains on member data. Since member identification constitutes the primary privacy risk in membership inference~\cite{carlini2022membership}, improvements on the member subset are of primary importance.

Overall, the results show that the proposed architectural interventions substantially reduce member-specific vulnerability -- the dominant privacy failure mode -- while largely preserving privacy on non-member data.

\begin{table}[t]
\centering
\scriptsize
\begin{tabular}{l|c}
\toprule
Model & Perplexity (Member) \\
\midrule
None        & 68.96 \\
Bottleneck Projection & 138.44 \\
NoNorm     & 52.99 \\
All        & \textbf{51.68} \\
\bottomrule
\end{tabular}
\caption{Perplexity on held-out training data points for DistilPythia student models under proposed architectural improvements. \textbf{Bolded} entries denote lower perplexity (better language-model utility).}
\label{tab:bottleneck-nonorm-ppl}
\end{table}

\subsection{Model Utility}

We evaluate model utility using perplexity on the member subset, as shown in Table~\ref{tab:bottleneck-nonorm-ppl}. \textbf{Bottleneck Projection} exhibits higher perplexity, reflecting the expected cost of reduced representational capacity. In contrast, \textbf{NoNorm} improves perplexity relative to the model with no privacy protection (\textbf{None}), and the combined model (\textbf{All}) achieves the lowest perplexity among all variants, demonstrating that \textbf{NoNorm} effectively mitigates the utility loss introduced by \textbf{Bottleneck Projection}.

Overall, the results demonstrate a favorable privacy-utility trade-off: while \textbf{Bottleneck Projection} alone reduces membership leakage (as illustrated in Table~\ref{tab:bottleneck-nonorm-acc}) at some cost to utility (as illustrated in Table~\ref{tab:bottleneck-nonorm-ppl}), combining \textbf{Bottleneck Projection} with \textbf{NoNorm} recovers model utility while maintaining a substantial reduction in membership inference success.

\section{Conclusion and Future Work}
\label{sec:conclusion}

In this paper, we study how knowledge distillation affects membership inference vulnerability in large language models. Across six teacher-student model pairs and six MIAs, we find that distillation does not reliably improve privacy: student models can match or exceed their teacher models in aggregate and member-specific attack success. This arises from mixed supervision during distillation, where alignment between teacher model predictions and ground-truth labels on vulnerable training data points induces overconfident student model behaviors.

We propose three mitigation strategies: restricting distillation to non-vulnerable data, introducing a low-dimensional \textbf{Bottleneck Projection}, and replacing layer normalization with \textbf{NoNorm}. While data restriction increases perplexity, the architectural modifications reduce member-specific MIA success without degrading utility.

Future work includes theoretical analyses of memorization under mixed supervision, adaptive privacy-utility trade-offs, integration with formal privacy mechanisms, and evaluation across broader model families and tasks.

\section*{Limitations}

This work opens several opportunities for further investigation. First, experiments are restricted to a subset of model families and teacher-student pairings (Pythia, Gemma 2, and Llama variants), which leaves open whether the observed amplification of member-specific leakage generalizes to other architectures and different distillation protocols. Second, evaluation focuses on a small collection of datasets (ArXiv and WikiMIA), so dataset characteristics (domain, language, multimodality) may affect memorization and attack efficacy. Third, the proposed mitigations are assessed empirically, and we do not combine them with mechanisms that provide formal privacy guarantees, such as DP-SGD. Addressing these issues will be important next steps to validate the practical utility of the proposed defenses.

\section*{AI-Generated Content Acknowledgement}
Language models were used to check for grammatical mistakes. Language models were also used to select appropriate wordings and improve sentence flow.


\bibliography{custom}

% \newpage
\appendix

\section{Optimal MIA Hyperparameters}
\label{app:hyperparam}

This appendix reports the optimal hyperparameters selected via cross-validation for the MIAs evaluated in this work, as described in Section~\ref{experimental setup}. We report the selected value of $k$ for Min-K\% and Min-K\%++, and the prefix length for ReCaLL in Table~\ref{tab:optimal-hyperparameters}. 

\begin{table}[t]
\centering
\tiny
\caption{Optimal hyperparameters selected via cross-validation for Min-K\%, Min-K\%++, and ReCaLL across all evaluated target models.}
\label{tab:optimal-hyperparameters}
\begin{tabular}{lccc}
\toprule
\textbf{Model} & \textbf{Min-K\% ($k$)} & \textbf{Min-K\%++ ($k$)} & \textbf{ReCaLL Prefix} \\
\midrule
Pythia & 0.10 & 0.50 & 7 \\
DistilPythia & 0.50 & 0.30 & 7 \\
Gemma 2 27B & 0.15 & 0.25 & 5 \\
Gemma 2 9B & 0.20 & 0.90 & 30 \\
Gemma 2 2B & 0.20 & 0.10 & 28 \\
Gemma 2 2B Distilled & 0.40 & 0.20 & 28 \\
Llama 3.1 8B & 0.70 & 0.10 & 20 \\
Llama 3.2 1B & 0.90 & 0.10 & 22 \\
Llama 3.2 3B & 0.90 & 0.10 & 22 \\
\bottomrule
\end{tabular}
\end{table}

\section{Ablation Study: Bottleneck Projection Dimensionality}
\label{app:ablation-bottleneck}

\begin{table}[t]
\centering
\scriptsize
\caption{Ablation over \textbf{Bottleneck Projection} dimension $B$, measured using MIA accuracies. Within each column, \textbf{bold} indicates the lowest accuracy (best for privacy) and \underline{underline} indicates the highest accuracy (worst for privacy) across different $B$ values.}
\label{tab:ablation-bottleneck-acc}
\begin{tabular}{l|cccccc}
\toprule
$B$ & ReCaLL & Loss & Zlib & Min-K & Min-K++ & Ref \\
\midrule
\multicolumn{7}{c}{\textbf{Evaluated on Member ($\mathcal{D}$)}} \\
\midrule
48  & 0.877 & 0.963 & \textbf{0.654} & \textbf{0.864} & 0.877 & 0.988 \\
96  & 0.864 & \textbf{0.914} & \textbf{0.654} & \underline{0.938} & \textbf{0.864} & \textbf{0.938} \\
192 & \underline{0.926} & \underline{0.975} & 0.679 & 0.889 & 0.914 & 0.988 \\
384 & \textbf{0.802} & 0.951 & 0.667 & 0.926 & 0.901 & 0.975 \\
768 & \underline{0.926} & \underline{0.975} & \underline{0.704} & 0.889 & \underline{0.938} & \underline{1.000} \\
\midrule
\multicolumn{7}{c}{\textbf{Evaluated on Non-member ($\mathcal{D}'$)}} \\
\midrule
48  & 0.889 & \textbf{0.951} & 1.000 & 0.951 & 0.938 & 0.988 \\
96  & \textbf{0.852} & \underline{0.975} & 1.000 & \textbf{0.852} & 0.938 & \underline{1.000} \\
192 & 0.926 & \textbf{0.951} & 1.000 & 0.951 & \textbf{0.914} & 0.988 \\
384 & 0.926 & \underline{0.975} & 1.000 & 0.951 & 0.926 & \underline{1.000} \\
768 & \underline{0.938} & 0.963 & 1.000 & \underline{0.963} & \underline{0.951} & 0.988 \\
\bottomrule
\end{tabular}
\end{table}

\begin{table}[ht]
\centering
\small
\caption{Perplexity on member subset for different \textbf{Bottleneck Projection} dimensions $B$. \underline{Underlined} values indicate highest perplexity (worst utility); \textbf{bold} values indicate lowest perplexity (best utility).}
\label{tab:ablation-bottleneck-ppl}
\begin{tabular}{l|c}
\toprule
$B$ & Perplexity (Member) \\
\midrule
48  & 200.17 \\
96  & \underline{218.67} \\
192 & 161.56 \\
384 & 138.44 \\
768 & \textbf{134.99} \\
\bottomrule
\end{tabular}
\end{table}

\paragraph{Setup.}
We evaluate the effect of the \textbf{Bottleneck Projection} dimensionality $B$ on MIA vulnerability and model utility. For this analysis, we fix all other training hyperparameters and train DistilPythia student variants for 30 epochs while varying $B\in\{48,96,192,384,768\}$. We report MIA accuracies for six representative attacks (ReCaLL, Loss, Zlib, Min-K\%, Min-K\%++, and Reference-model) and perplexity measured on the member subset.

\paragraph{Findings.}

% \begin{table}[t]
% \centering
% \scriptsize
% \caption{Ablation over Bottleneck Projection dimension $B$, measured using MIA accuracies. Within each column, the \underline{lowest} accuracy (best for privacy) and the \textbf{highest} accuracy (worst for privacy) across $B$ are highlighted.}
% \label{tab:ablation-bottleneck-acc}
% \begin{tabular}{l|cccccc}
% \toprule
% $B$ & ReCaLL & Loss & Zlib & Min-K & Min-K++ & Ref \\
% \midrule
% \multicolumn{7}{c}{\textbf{Evaluated on Member ($\mathcal{D}$)}} \\
% \midrule
% 48  & 0.877 & 0.963 & \underline{0.654} & \underline{0.864} & 0.877 & 0.988 \\
% 96  & 0.864 & \underline{0.914} & \underline{0.654} & \textbf{0.938} & \underline{0.864} & \underline{0.938} \\
% 192 & \textbf{0.926} & \textbf{0.975} & 0.679 & 0.889 & 0.914 & 0.988 \\
% 384 & \underline{0.802} & 0.951 & 0.667 & 0.926 & 0.901 & 0.975 \\
% 768 & \textbf{0.926} & \textbf{0.975} & \textbf{0.704} & 0.889 & \textbf{0.938} & \textbf{1.000} \\
% \midrule
% \multicolumn{7}{c}{\textbf{Evaluated on Non-member ($\mathcal{D}'$)}} \\
% \midrule
% 48  & 0.889 & \underline{0.951} & \textbf{1.000} & 0.951 & 0.938 & 0.988 \\
% 96  & \underline{0.852} & \textbf{0.975} & \textbf{1.000} & \underline{0.852} & 0.938 & \textbf{1.000} \\
% 192 & 0.926 & \underline{0.951} & \textbf{1.000} & 0.951 & \underline{0.914} & 0.988 \\
% 384 & 0.926 & \textbf{0.975} & \textbf{1.000} & 0.951 & 0.926 & \textbf{1.000} \\
% 768 & \textbf{0.938} & 0.963 & \textbf{1.000} & \textbf{0.963} & \textbf{0.951} & 0.988 \\
% \bottomrule
% \end{tabular}
% \end{table}

The privacy attack results are shown in Table~\ref{tab:ablation-bottleneck-acc}. For the member side, smaller \textbf{Bottleneck Projection} dimensions often reduce MIA accuracy. In particular, $B=768$ attains the highest MIA accuracy across 5 of the 6 MIA methods, and there is a general decreasing trend in MIA accuracy as $B$ decreases. For the non-member case, we also observe that $B=768$ achieves the highest MIA accuracy across 4 of the 6 MIA methods, and a similar decreasing trend as before. Therefore, we conclude that a bottleneck indeed has the ability to limit memorization and thus vulnerability to MIA.

The perplexity results are shown in Table~\ref{tab:ablation-bottleneck-ppl}. Perplexity on members generally improves as $B$ increases; the largest $B$ tested ($B=768$) attains the lowest perplexity (best utility, 134.99), whereas $B=96$ produced the worst perplexity (218.67). Intermediate values ($B=192$ and $B=384$) provide a compromise between privacy and utility.

\end{document}